\documentclass{article}

\AtBeginDocument{%
  }

\usepackage[utf8]{inputenc} 
\usepackage{kotex} 

\usepackage{arxiv}

\usepackage[colorinlistoftodos,prependcaption,textsize=small]{todonotes}
\usepackage{caption}
\usepackage{subcaption}
\usepackage{amsmath}
\usepackage[capitalise, noabbrev, nameinlink]{cleveref}
\usepackage{algorithm} 
\usepackage{algpseudocode}
\usepackage{multirow}
\usepackage{makecell}
\usepackage[english]{babel}

\definecolor{red}{RGB}{255, 0, 0}
\definecolor{gre}{RGB}{0, 255, 0}
\definecolor{blu}{RGB}{0, 0, 255}
\definecolor{pur}{RGB}{255, 0, 255}
\newcommand{\red}[1]{\textcolor{red}{#1}}

\newcommand{\nnlm}{\textbf{NNLM}}

\newcommand{\et}{\textit{et al.}}
\newcommand{\ie}{\textit{i.e.}}
\newcommand{\lol}{\textit{League of Legends}}
\newcommand{\dota}{\textit{Dota 2}}
\newcommand{\hok}{\textit{Honor of the Kings}}

\newcommand{\bb}[1]{\textbf{#1}}
\newcommand{\ii}[1]{\textit{#1}}

\begin{document}
\title{Action2Score: An Embedding approach to Score Player Action}

\author{
  Junho Jang \\
  Korea University \\
  Seoul, Republic of Korea \\
  \texttt{hkonly@korea.ac.kr} \\
  \And
  Ji Young Woo \\
  Soonchunhyang University \\
  Asan, Republic of Korea \\
  \texttt{jywoo@sch.ac.kr}\\
  \And
  Huy Kang Kim \\
  Korea University \\
  Seoul, Republic of Korea \\
  \texttt{cenda@korea.ac.kr} \\
}




\maketitle

\begin{abstract}
Multiplayer Online Battle Arena (MOBA) is one of the most successful game genres. MOBA games such as League of Legends have competitive environments where players race for their rank. In most MOBA games, a player’s rank is determined by the match result (win or lose). It seems natural because of the nature of team play, but in some sense, it is unfair because the players who put a lot of effort lose their rank just in case of loss and some players even get free-ride on teammates’ efforts in case of a win. To reduce the side-effects of the team-based ranking system and evaluate a player’s performance impartially, we propose a novel embedding model that converts a player’s actions into quantitative scores based on the actions’ respective contribution to the team’s victory. Our model is built using a sequence-based deep learning model with a novel loss function working on the team match. 
We showed that our model can evaluate a player’s individual performance fairly and analyze the contributions of the player’s respective actions.
\end{abstract}



\section{Introduction}\label{sec:intro}
`Multiplayer Online Battle Arena (\bb{MOBA})' is a game genre that represents team competition strategy games where two five-player teams face off to destroy each other team's base. The MOBA is one of the most successful games genres nowadays. Especially, in the League of Legends, one of the MOBA games, more than 70 million people watched its world championship in 2021 \cite{championship_viewer}. Also, about 180 million players are enjoying the game in 2022 \cite{active_players}. Most MOBA games have a `Rank' system that evaluates their players and groups them at a similar level. Players can achieve a higher Rank by increasing their `Match Making Rating (\bb{MMR}).' The MMR is designed to match players who have similar levels. While most game companies do not open the exact formulas of the MMR systems in public, game communities surmise that the games such as \lol, \dota, and \ii{Honor of Kings} use revised versions of the Elo rating system \cite{elo_lolboosts, elo_dota2freaks} used in chess leagues \cite{wiki_elo}. In the Elo rating mechanism, a player's MMR increases or decreases by a game match's result (win/lose). The system creates a competitive environment and gives players a sense of achievement when they reach a higher rank.

It is reasonable that a team's achievement (win/lose) directly affects a player's MMR since MOBA games are team-playing games. In many cases, players of the lost team performed relatively less-valuable acts than the won team players. However, some players perform remarkably although the team is losing. On the other hand, some players take advantage of teammates' performance without their skill and effort. Therefore, estimating individual performances in team-play matches is also a challenging and attractive topic in the sports domain.

In addition, occasionally, the competitive environment and the fall of a player's rank trigger undesired side-effects such as losing motivation and toxic behaviors. For example, players can encounter an incompetent player (intentionally or unintentionally) as a teammate. In that case, the players usually exercise `Peer pressure' \cite{heneman1995balancing} towards the unskilled player because they worry about their rank falling if they lose. Peer pressure strategy sometimes works fine. However, it generally brings rage and offensive reactions to the peer-pressured player \cite{kou2014playing}. Hence, reflecting individual contributions to the team's achievement to MMR can relieve the side-effects of a current team-based reward system.

Many game communities use metrics to evaluate a player's individual performance (See \cref{sec:metrics}). However, the metrics have limitations; they evaluate a player's contribution for a whole match, while evaluations of the respective actions of the player are absent. Also, the metrics do not consider the context of when an indicator transition occurs. For example, a player can die for a strategic purpose, such as delaying enemy march until teammates are ready to fight or sacrifice themselves to save teammates from a massacre. Therefore, there are some outlier players that are underestimated or overestimated by those metrics.

We propose \bb{an embedding approach that embeds a player's respective actions to quantified scores.} Our approach uses combinations of recurrent neural networks and multi-layered perceptrons and is also inspired by the concept of the neural network language model (NNLM), a word-embedding approach. The main contributions derived from our approach are below:

\begin{enumerate}
    \item The proposed model allows the game system to quantify an individual player's contribution to the team's victory to alleviate the disadvantage of the current MMR system.
    \item The model can discover and re-evaluate players who are over/underestimated by common performance metrics.
    \item The model can guide a player to improve their skills by debriefing a match analysis with the quantified scores of their respective actions.
\end{enumerate}

This paper is organized as follows. First, we describe the basic rules of \lol~and introduce common metrics to evaluate a player's individual performance in \cref{sec:lol}. Next, \cref{sec:related_works} introduces previous studies to evaluate a player's contribution and provides background on the NNLM and RNN that underlie our model. Then, \cref{sec:model} and \cref{sec:dataset} present our proposed model and the dataset we used to train and validate our model, respectively. Next, we evaluate our model's performance by comparing it with common metrics as baselines and by more detailed analysis, in \cref{sec:experiment}. In \cref{sec:discussion}, In \cref{sec:discussion}, we review the experiment result and discuss the limitations and strengths of this study. Finally, we conclude the paper by noting the contribution of this paper and the prospect of usage of the model we proposed, In \cref{sec:conclusion}.

\section{League of Legends}\label{sec:lol}
\begin{figure}[ht]
    \centering
    \includegraphics[width=0.9\linewidth]{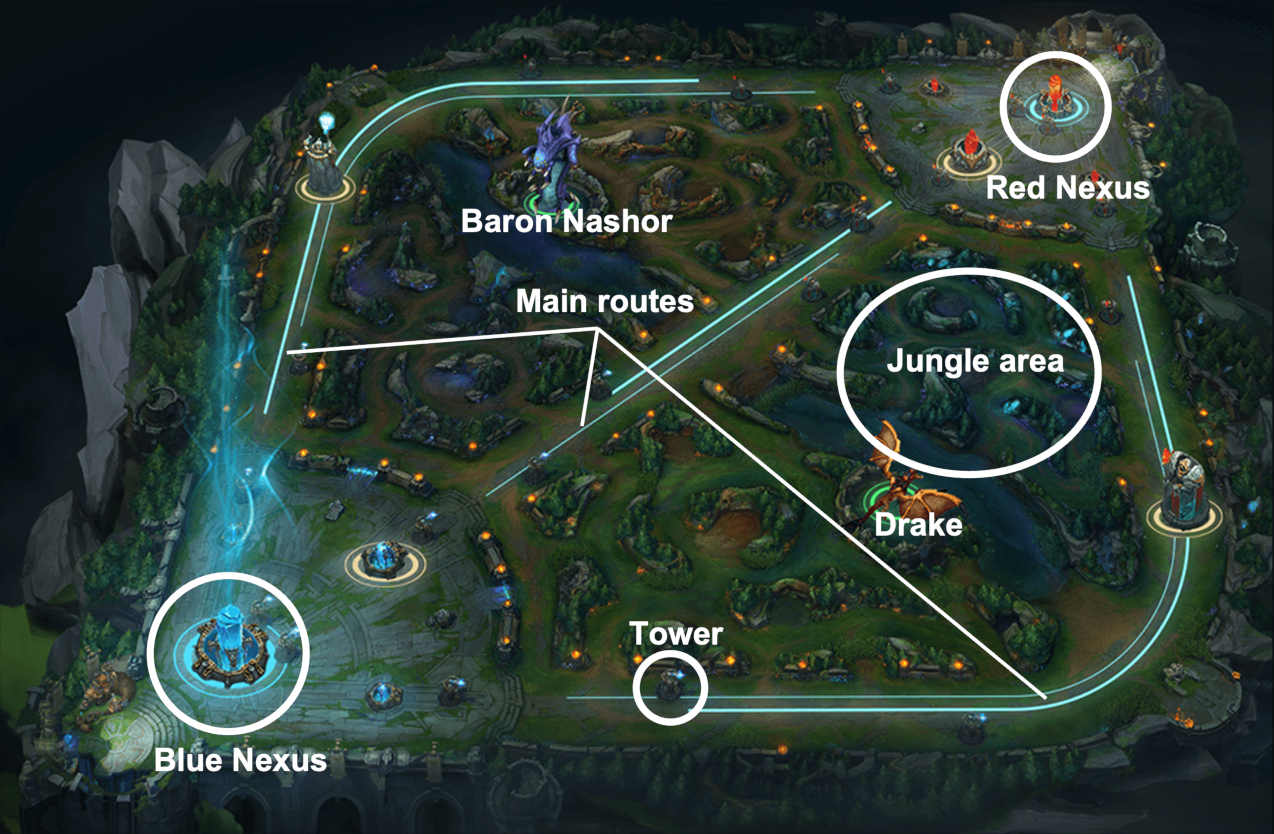}
    \caption{Quarter view of map of \lol.}
    \label{fig:map}
    \vspace{10pt}
\end{figure}

\lol~ is one of the MOBA games that many players in the world enjoy every day. The Primary system of \lol~ is similar to other MOBA games; Ten players meet in a single match and form teams of five players each. There are two bases called \bb{Nexus} for both teams on either side of the map diagonally (See \cref{fig:map} \cite{lol_official}). The lower-left corner base team is \bb{blue}, and the opposite is \bb{red}. The ultimate goal of both teams is to destroy the opponent's Nexus while protecting their own. There are three main roads to reach the opposing bases, and \bb{defense towers} belonging to both teams prevent opposing players from accessing their Nexus.

\subsection{Champions}
In a match, a player chooses a playable character called \bb{champion}. The game has over 140 champions, and a champion belongs to one or more of the following \bb{roles}: \ii{Assassins}, \ii{Fighters}, \ii{Mages}, \ii{Marksmen}, \ii{Supports}, \ii{Tanks}. Assassins are killers with excellent damaging ability but are vulnerable to being killed because of their low durability. On the other hand, Fighters have both good damaging ability and survivability. They commit in short-ranged combat. Tankers take and endure the enemy's attack instead of teammates with extraordinary durability. Also, Marksmen make long-ranged attacks behind other teammates. Mages are the champions who damage, debuff (make a target weaker) or disturb the enemy with their magic. Last, Supports help teammates with various skills, such as healing and buff. These are the typical character classes of most role-based games like RPGs and MOBAs.

\subsection{Lanes}
\bb{Lane} means the three-main routes of a map: \ii{Top}, \ii{Mid}, and \ii{Bottom}. Each lane indicates the upper, middle, and lower roads, respectively.
Meanwhile, there is another terminology called \bb{Position}, which denotes players' roles in their team. The game officially designed five positions, \ii{Top}, \ii{Mid}, \ii{Bottom}, \ii{Jungle}, and \ii{Support}. The players who take Top, Mid, and Bottom positions are responsible for defending the enemy's march in their corresponding lane and pushing the line of combat in the direction of the opponent's base. On the other hand, the Jungle player roams in the jungle area while hunting monsters and joins the engagement in the lanes. Occasionally they assassinate enemy champions with surprise. Also, the Support player aids other teammates with their assisting skills such as buff or healing. They usually company the Bottom player. There are three duplicating terms between Land and Position--Top, Mid, and Bottom. Therefore, the game community often uses the terminology Lane and Position interchangeably. In this paper, we use the term \bb{Lane} rather than Position because we use the term Position for another meaning.

\subsection{Minions and Monsters}
\bb{Minions} are NPC soldiers that belong to one of the two teams. Both teams' minions are continuously generated from their base, and they charge to the opponent's base, following the three main roads and attacking enemy characters and structures. On the other hand, \bb{Monsters} are NPCs that do not belong to any team. Players can collect gold and experience from slaying monsters in the jungle areas. Especially, \ii{Baron Nashor} and \ii{Drake} are called the \bb{Elite monsters} that give significant bonuses (such as buffs) to the team which hunts them.

\subsection{Kill, Support, and Death}
While struggling to achieve the team goal and intervening enemy, players can kill the opponent players and aid teammates. The game records a player who gave the last damage to the victim as the killer. On the other hand, a player who damaged the victim in the last 10 seconds before the victim died gets the assist record.
Meanwhile, the players whom the enemy killed are not washed out from the game. Instead, they rebirth after certain times elapsed from the death. Therefore, victims can continue the match; however, their team will be at a disadvantage of a troops shortage while the victims wait for rebirth.

\subsection{Gold, Item, and Experience}
\bb{Gold} is the currency of the \lol~ that players can use to buy items. Players can collect gold from several actions, such as killing enemy champions and minions, hunting monsters, and destroying enemy structures. The items provide tactical functionality like healing the owner and stuns the enemy.
Furthermore, Players also get \bb{experience} from the actions mentioned above, and they can level up with the cumulated experience; it gives the players enhanced ability and unlock additional skills.

\subsection{Common metrics to evaluate individual performance}\label{sec:metrics}
As mentioned in \cref{sec:intro}, many game communities commonly reference some metrics to estimate a player's performance (hereinafter \bb{common metrics}). The most common metrics are the Kill-Death-Assists ratio (KDA), Gold, and Creep score. \bb{KDA} represents the ratio that a player recorded how many kills and assists compared to their death. \bb{Gold} represents the amount of gold that a player earned in a match. \bb{Creep score} denotes the number of monsters and minions that a player hunted. A player records 1 creep score for hunting a minion or a regular monster and records 4 creep scores for hunting a giant monster.

\section{Related Works}\label{sec:related_works}
\subsection{Win prediction}
Predicting the outcome of a match is one of the most actively researched areas. Of course, the topic is interesting itself; however, it is also momentous because many studies that estimate player contribution use the prediction results.

Most studies use one or both of the two types of information to predict outcomes: pre-match information and mid-match information. Pre-match information denotes information collected before the match starts, such as champions that players selected and players' gameplay history. On the other hand, mid/post-match information represents the information that occurred during a match, including KDA, gold, and creep score.

The eight papers among the win prediction studies we surveyed use only pre-match information. Conley \et~applied players' choice of heroes to logistic regression and the K-nearest neighbors model to predict match outcome and recommend the player choose a better hero to win. The logistic regression model shows up to 71\% of prediction accuracy \cite{conley2013does}. Kalyanaraman \et~introduced a regression model that uses combinations of heroes in \dota. Their model shows 69.42\% prediction accuracy \cite{kalyanaraman2014win}. Song \et~also use hero draft data of \dota~in the logistic regression model to predict match outcome. They achieved a maximum of 58\% accuracy with their model. To improve the model's performance, the authors claim they added some hero combo features, and the model recorded 61\% accuracy with the features. However, the authors do not explain the details of the hero combo features \cite{song2015predicting}. Semenov \et~also proposed a prediction model for \dota~that receives heroes draft as input based on Na\"ive Bayes, Logistic regression, and Gradient boosted decision tree. Their model predicted games' results with 70.6\%, 67\% and 66\% accuracy for average, high skilled, very high skilled player games, respectively \cite{semenov2016performance}. Hanke \et~suggested an MLP model that uses hero draft data as inputs for prediction in \dota. Their model displayed 88.63\% prediction accuracy. They also proposed a hero recommendation system for players based on the frequent itemset mining with the apriori algorithm. The win percentile of the recommended hero combinations was 74.9\% \cite{hanke2017recommender}. Andono \et~ use Na\"ive Bayes model with Adaboost algorithm to predict match outcomes of \dota. The model records 80\% prediction accuracy with metadata of heroes, such as heroes' types and attributes. However, the dataset used for the experiment only includes matches with one human player and four AI players for each team; therefore, it is difficult to generalize the results of this study to common matches that consist of 10 human players \cite{andono2017dota}. Do \et~apply the players' records on champions, such as win rate with a champion and the total number of the played game with the champion, as a feature to their deep neural network model. They trained their model with 5,000 \lol~match data and achieved 75.1\% of prediction accuracy \cite{do2021using}. Lee \et~ proposed a personalized champion recommendation system for \lol~and \dota~based on the win prediction rate of a player's preferred champions. The authors divided the win prediction model into player-level and match-level embedding networks. The player-level embedding networks transform a player stat into a vector; the vector is input to match-level embedding networks that use champions and their role data with the player's stats to predict a match outcome. The model recorded its highest prediction accuracy on \lol~as 55.35\% and 57.55\% on \dota~\cite{lee2022draftrec}.

Six studies apply mid or post-match information to predict the outcome of a match. Rioult \et~proposed a binary classification model that uses topological clues, and their model achieved 85\% precision and 90\% recall. However, the number of replay data they used is insufficient as the authors recognize \cite{rioult2014mining}. Kinkade \et~ introduced two approaches that respectively use pre-match and post-match information. First, they used hero matchup and heroes' synergy/countering data for the pre-match information. Their pre-match model indicated up to 73\% accuracy. On the other hand, their post-match model uses gold, experience, and kill numbers that are cumulated per minute. The post-match model shows a maximum of 99.81\% of accuracy \cite{kinkade2015dota}. Hodge \et~proposed a real-time prediction model for professional matches of {\itshape Dota 2}, using Logistic Regression, Random Forest, LightGBM, respectively \cite{hodge2019win}. They achieved the accuracy to 74.59\% with recorded game data; moreover, their model hits 85\% accuracy when predicts the outcome at 5-minute game time elapsed with live data from \textit{the ESL One Hamburg 2017 Dota 2 league} . Yang \et~attempted not just to predict the outcome of matches but also to interpret which features were essential to derive the outcome, using 184 thousand match data of \hok~\cite{yang2022interpretable}. Their model consists of two stages, spatial and temporal. First, six features, including pre-match and mid-match data, are processed in the spatial stage with logistic regression and feed-forward networks. Then temporal stage draws the final prediction from the results of the spatial stage. TSSTN shows 54.6\% accurate prediction at the start of a match and a maximum of 78.5\% accuracy when 10 minutes elapsed. Another research by Yang \et~ uses mid-match statistical information and the occurrence of champion and boss monster kill events with a two-directional LSTM/transformer model and fully connected neural networks to predict the match outcome and the killer and victim of the next champion and boss monster kill events \cite{yang2022predicting}. Their model performs 70.8\% of win prediction accuracy and predicts the killer of the next champion and boss monster with a maximum of 94.4\% and 28.1\% accuracy, respectively. Zhao \et~proposed a real-time prediction model, Winning Tracker \cite{zhao2022winning}. They reconstructed players and towers data of \lol~matches into confrontation and individual movement information to predict the match outcome and the next tower destruction event. The model achieved up to 0.901 to predict tower destruction and 0.889 on match outcome prediction, on F1-score.

\subsection{Player contribution}
Estimating the performance of individual players is an attractive topic in the sports analysis domain, including e-sports. The studies that attempted to evaluate the contributions of players of MOBA games share a similar goal with this study. Suznjevic \et~proposed Application Context-Aware Rating algorIthm (\bb{ACARI}), which is an algorithm to adjust the MMR system on \ii{Heroes of Newerth} (HoN) \cite{suznjevic2015application}. They created hero vectors representing how much a \bb{hero} (playable character of HoN) fits various \bb{roles} that a player can take in a match. Then they choose parameters related to a player's performance; the proportion of a player's parameter among the teammates is the player's contribution toward the team's achievement. The authors also modified the contributions with two weights. The one weight is defined with correlations between a hero and each parameter, derived from the domain knowledge. The other weight comes from the parameter percentile on the 10000 matches statistics. Finally, they adjusted the increasing/decreasing MMR with estimated individual performances after a match. Cavandenti \et~proposed a pattern analysis model to help novice players to improve their skills by analyzing their behaviors in \dota~\cite{cavadenti2016did}. They mined behavioral patterns of skillful players and compared them to novice players' acts. Their anomaly detection-like model is helpful to analyze whether a novice player's action is closed to the standard play style of skillful players. However, their model cannot explain whether the skillful players' play patterns are beneficial to lead the team's victory or not. That makes the limitation that their model is hard to estimate players' individual performances. Sapienza \et~ considered only KDA as a performance indicator for an individual player \cite{sapienza2018individual}. The authors reveal that a player's win rate and performance tend to decrease when a player continuously plays the game. Jiang \et~classified League of Legends players into four categories from 148,000 match data using the concept of diversity and conformity; maverick, generalist, specialist, and niche players. \cite{jiang2021wide}. They used players' champion selection and movement to classify them. The authors also analyzed players' performance in each category by KDA, the approximate value of MMR, and the win rate. Ramler \et~ conducted an interesting analysis of the existence of performance differences between genders of League of Legends champions \cite{ramler2021investigating}. The indicators that the study used to quantify a champion's performance are the end-game statistical data such as total minion kills, gold, and KDA. Maymin distinguished the Smart kill and the Useless death using win probability transition of before and after a killing event \cite{maymin2021smart}. The former means the kill action that increases the chance of the team's victory, and the latter is the death that decreases the probability of winning. They estimated middle-of-game win probability using a logistic regression model; the input features were champion kill count, the number of remaining towers, elite monster kill count, elapsed game time in minutes. Also, they analyzed additional features such as gold and survivability that how to affect those features toward the team's winning probability. Finally, Damediuk attempted to set a performance index to quantify individual players' performances with \dota match data \cite{demediuk2021performance}. They used end-game data such as XP, LEVEL, and KDA; and calculated the weighted sum of the features according to the player's playstyle, of which the authors classified a player into one of 10 types.

As reviewed above, many studies attempt to assess players' contributions. However, most studies use result-oriented measurements; and they do not try to assess the respective actions of players, except in a few studies such as Maymin's study. On the other hand, We intended to study the areas that were less explored by former studies; in this study, we attempt to assess the contributions of individual actions to a team's victory.

\subsection{Word embedding models}
\begin{figure}[ht]
    \centering
    \includegraphics[width=0.9\linewidth]{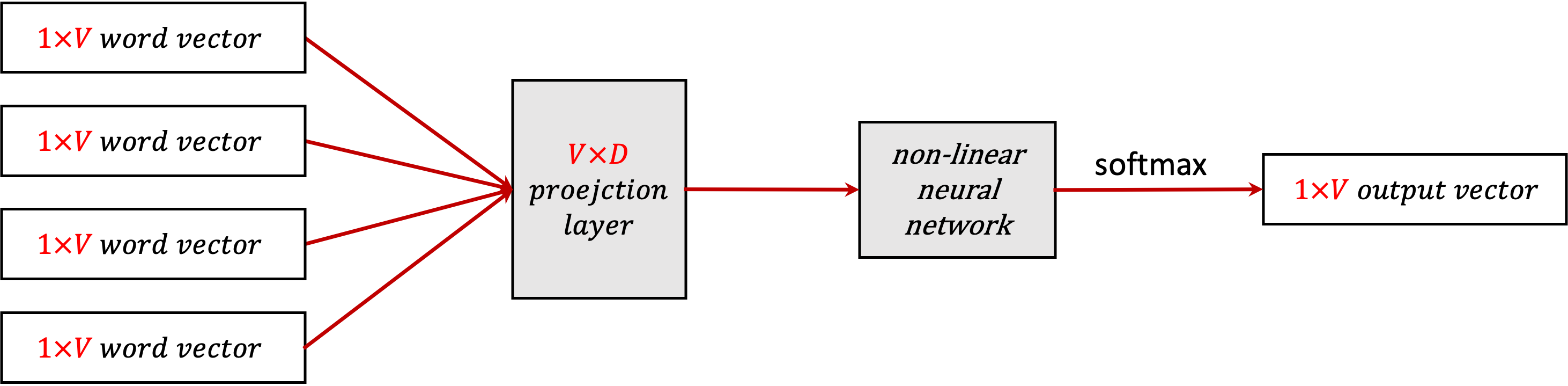}
    \caption{Basic process of NNLM}
    \label{fig:nnlm}
    \vspace{10pt}
\end{figure}

Word embedding is the conversion of text words into N-dimensional dense vectors. Embedded vectors make the model can process words and make it possible to represent relations between words mathematically. Neural Network Language Model (\nnlm) is an word embedding model that uses feed-forward neural network \cite{nnlm}. \nnlm~model consists of input, projection layer, hidden layer, and output. The model is trained to predict a next-coming word when it takes a $1 \times V$ dimensional words sequence in the context length, $N$, as an input. The projection layer is a $V \times D$ dimensional lookup table where each vector row is matched 1:1 with $V$ words in the vocabulary. The vectors in the projection layer are randomly initialized, and they are transformed into embedded word vectors that represent V words as training proceeds. \cref{fig:nnlm} shows the basic process of \nnlm.

\subsection{Recurrent neural network}
\begin{figure}[ht]
    \vspace{7pt}
    \centering
    \begin{subfigure}[b]{0.15\linewidth}
        \includegraphics[width=\linewidth, height=1.8\linewidth]{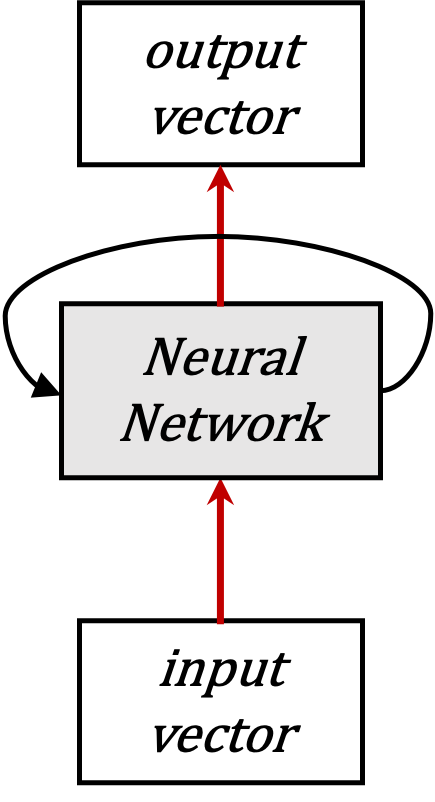}
        \caption{standard figure}
        \label{fig:rnn_std}
    \end{subfigure}
    \hfill
    \begin{subfigure}[b]{0.15\linewidth}
        \includegraphics[width=\linewidth, height=1.8\linewidth]{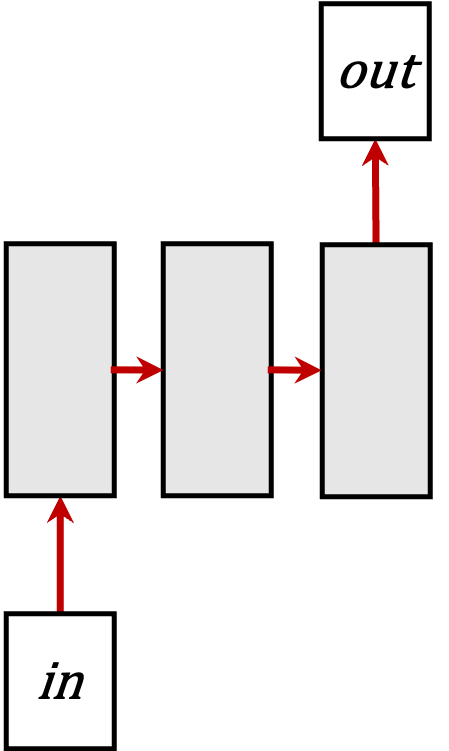}
        \caption{one-to-one}
        \label{fig:rnn_1to1}
    \end{subfigure}
    \hfill
    \begin{subfigure}[b]{0.15\linewidth}
        \includegraphics[width=\linewidth, height=1.8\linewidth]{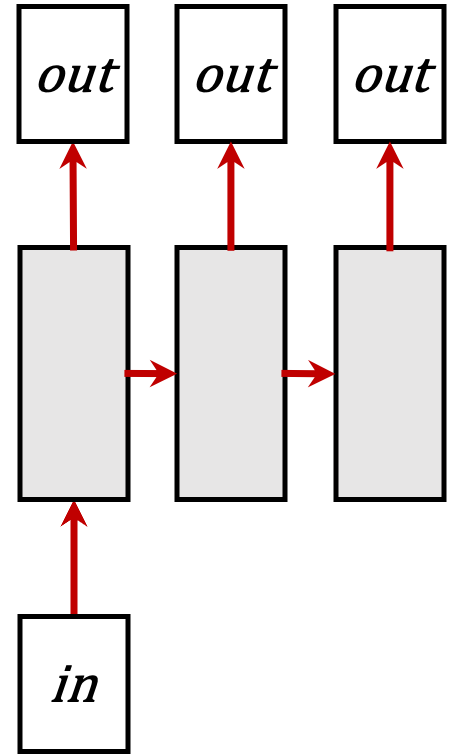}
        \caption{one-to-many}
        \label{fig:rnn_1ton}
    \end{subfigure}
    \hfill
    \begin{subfigure}[b]{0.15\linewidth}
        \includegraphics[width=\linewidth, height=1.8\linewidth]{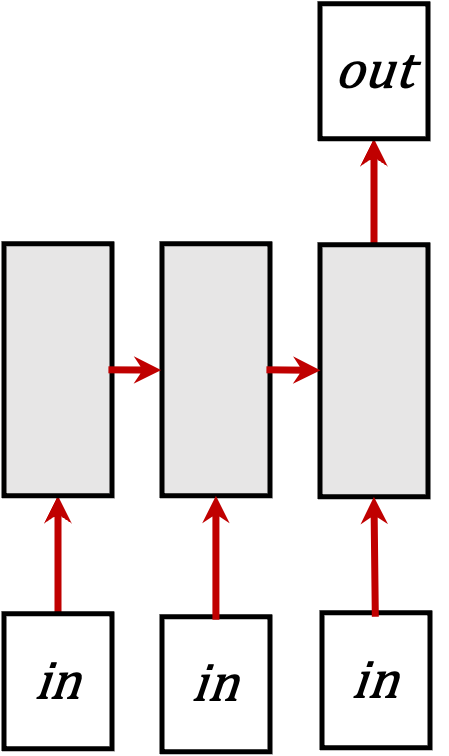}
        \caption{many-to-one}
        \label{fig:rnn_nto1}
    \end{subfigure}
    \hfill
    \begin{subfigure}[b]{0.15\linewidth}
        \includegraphics[width=\linewidth, height=1.8\linewidth]{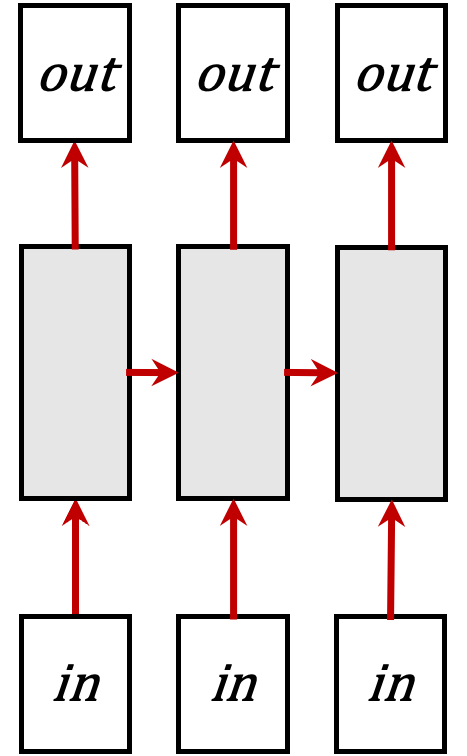}
        \caption{many-to-many}
        \label{fig:rnn_nton}
    \end{subfigure}
    \caption{Representative Types of RNN}
    \label{fig:rnn}
\end{figure}

A Recurrent Neural Network (RNN) is a type of neural network model for dealing with sequence data in which the outcome of one time point influences the outcome of the next.
In the RNN model, a neural network at time $t$ takes sequence data of $t$ and the neural network's output at $t-1$ as inputs. Basically, there are four types of RNN models; \ii{one-to-one}, \ii{one-to-many}, \ii{many-to-one}, \ii{many-to-many}. \cref{fig:rnn} shows the standard figure and four types of RNN.

Long Short-Term Memory (\bb{LSTM}) is an improved model that expiates the weaknesses of the standard RNN model \cite{lstm}. In the standard RNN model, information of previous steps gradually disappears if the information's weight is too small. In opposite, if the weight is too huge, previous information's influence becomes excessively strong. LSTM overcomes it using cell states, in which three gates determine whether the previous time point's data will be dropped or remembered.

Gated Recurrent Unit (\bb{GRU}) is a modified type of LSTM that uses only two gates \cite{gru}. By reducing the number of gates in a NN cell, GRU aims to decrease computational workloads to train.

\section{Proposed model}\label{sec:model}
In this paper, we propose an action-embedding model that converts a player's actions into quantitative scores. We borrow NNLM's word embedding idea. The difference between our model and NNLM is the presence or absence of a projection layer. The word embedding approach has a 1:1 matching relation between words and embedded vectors; therefore, NNLM has a projection layer and trains the layer's value. However, the relationship between a player's action types and the score is not one-to-one. For example, when a player kills an enemy's champion, the action is evaluated variously by contexts such as the timing and the victim's remaining health. Because of this reason, our model does not use a projection layer. Instead, our model aims to obtain parameters of neural networks that output adequate scores when the action features are given. \ie, the neural network's parameters of our model and projection layer of NNLM take a similar role.

Meanwhile, a player's action has causal relationships with the previous or the following action. So we can evaluate the cause action through the resulting action. For example, if a player killed an enemy champion after purchasing an item, the act of purchasing an item was likely beneficial. We combined the score-embedding model with the RNN model to reflect the causal relationship between actions. Moreover, different from regular RNN models, we implemented our model to take player actions in a time-reversed sequence to analyze the previous action through the following action. 

Our model consists of two main parts. One converts player action sequences into scores, and the other discerns the winner by comparing the two teams' scores and calculates the loss to execute backpropagation. \cref{sec:rnnslp} and \cref{sec:dep} describe the details of each part of the model.

\subsection{RNN and single layer perceptron part}\label{sec:rnnslp}
\begin{figure}[ht]
    \centering
    \vspace{7pt}
    \includegraphics[width=0.8\linewidth]{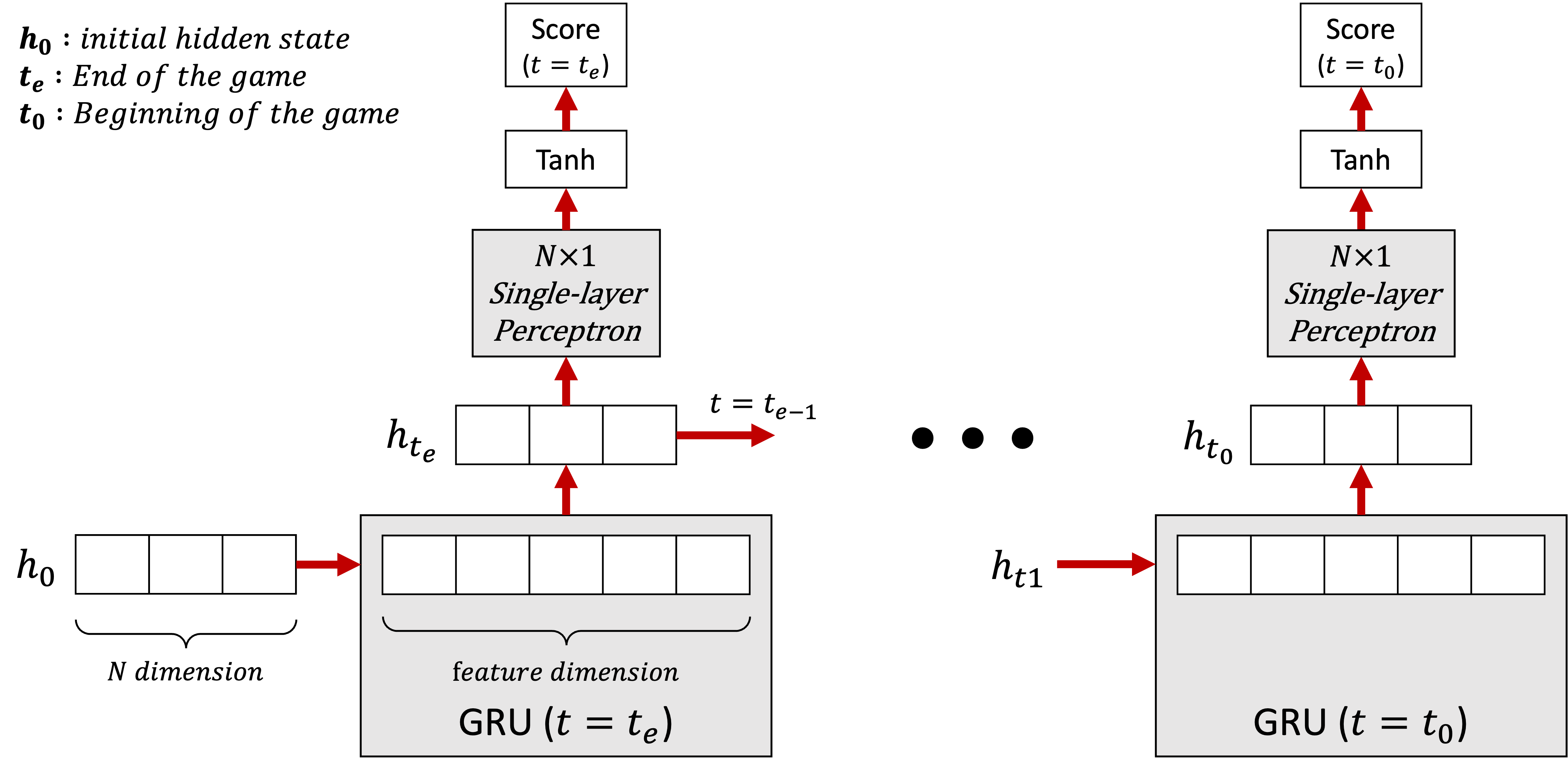}
    \caption{basic process of GRU-SLP part}
    \label{fig:gru+slp}
\end{figure}

The first part of our model consists of RNN and Single Layer Perceptron (\bb{SLP}). We used GRU for RNN. When a player's action sequence is input to GRU, the output (hidden state) is input to the next GRU and corresponding SLP. SLPs output a 1-dimensional vector respectively when taking N-dimensional hidden states of each GRU. Finally, each output of SLPs becomes scores between -1 to 1 real number through the hyperbolic tangent (\textit{tanh}) function. An action converted into a positive score represents beneficial action to win; the opposite is a harmful action that drove defeat. \cref{fig:gru+slp} depicts the operation of a GRU-SLP combined model.

\subsection{Discernment and evaluation part}\label{sec:dep}

As described above, the Discernment and Evaluation Part (hereinafter DEP) discerns the winner by comparing the two teams' scores and computes the loss to conduct backpropagation. After obtaining ten players' scores, DEP adds up blue and red team players' scores, respectively. DEP discerns which team is the winner through both teams' added-up scores. Also, it calculates loss using scores and discernment. We used two pairs of discernment method-loss function. The details of the sets are explained in \cref{sec:loss_function}.

Ten GRU-SLPs' parameter values are updated after backpropagation. In \cref{sec:overall} we describe the overall training process.

\subsection{Discernment methods and loss function pairs}\label{sec:loss_function}

As mentioned above, the goal is training our model to output scores that reflect the matches' outcomes properly. Our study stands based on the assumption that a team that acts on profitable actions more would win. Therefore, if a model gave a higher total score to the won team than the lost team, then the score reflected the match result well. On the other hand, if the lost team's total score is higher than the won team's, then the model misestimated the values of actions in the match.

We defined two discernment method-loss function pairs to train our model; \bb{Confidence--Cross Entropy} and \bb{Deterministic--ReLU}.

\subsubsection{\bb{Confidence and cross-entropy loss}}
In Confidence--Cross Entropy pair, DEP calculates how confident a team given a higher score is the actual winner. For example, there are two matches; On the first, the blue team got 20 points of the score, and the reds got 5. On the other, the blues got 10 points, and the reds got 7. In this case, DEP is more confident in the analysis result of the first match than the second. Through a softmax, the confidence is expressed as a probability in a real number between 0 - 1 function. \cref{eq:confidence} shows how DEP calculates the confidence $c(T)$. $T$ is a team among both teams, and $S_T$ is the team's total score. $S_B$ and $S_R$ represents the total score of the blue and red team, respectively.

Cross entropy is a function that outputs how different between two given probability distributions $P$ and $Q$ using information entropy theory. The output of the cross-entropy function represents the weighted arithmetic mean of the information content of when an event with information content according to probability distribution Q occurs according to probability distribution P. Many classification models use the function as the loss function to compare output probability distribution with the ground truth. We use cross-entropy to calculate the error between our model's analysis and the actual result. \cref{eq:crossentropy} displays how our model adopts the cross-entropy loss function. The values of $q(T_n)$ are determined by which team is the actual winner. If blue team is the winner, the values are: $q(T_{blue})=1.0$, $q(T_{red})=0.0$; while $q(T_{blue})=0.0$, $q(T_{red})=1.0$ if red team is the winner.

We used binary cross-entropy (BCE) loss to simplify codes and reduce computing workloads instead of cross-entropy loss on the implementation level. For BCE loss, confidence is a degree of certainty that the blue team is the actual winner. The calculation of modified confidence $c'(T)$ and BCE loss is shown in \cref{eq:s_confidence} and \cref{eq:bce}.

\begin{equation}\label{eq:confidence}
    \vspace{3pt}
    c(T) = \frac{e^{S_T}}{e^{S_B} + e^{S_R}}
    \vspace{3pt}
\end{equation}

\begin{equation}\label{eq:crossentropy}
    \vspace{3pt}
    CE\ loss = -\sum_{n}{q(T_n)\ log\ c(T_n)}
    \vspace{3pt}
\end{equation}

\begin{equation}\label{eq:s_confidence}
    \vspace{3pt}
    c'(T) = \frac{e^{S_B-S_R}}{e^{S_B-S_R} + 1}
    \vspace{3pt}
\end{equation}

\begin{equation}\label{eq:bce}
    \vspace{3pt}
    BCE\ loss = -[q(T_B)\ log\ c'(T_B)+(1-q(T_B))\ log\ (1-c'(T_B))]
\end{equation}

\subsubsection{\bb{Deterministic and ReLU loss}}
Sometimes a team can overwhelm their opponent and win a match with a massive performance gap. However, in many sports games that two or more teams compete, it is not uncommon for one team to win by a very narrow margin of performance. In this case, the lost team can get a high score from the model as much as the won team, and that is reasonable. However, with Confidence--Cross Entropy pair, the loss is very high when the model scores both teams with narrow differences, even if the discernment of the winner is correct.

To alleviate this problem, we designed Deterministic-ReLU pair. With this pair, DEP does not calculate the confidence. Instead, DEP discerns the winner only by comparing the total scores of both teams. Therefore, the discernment for the winner is expressed in deterministic.

ReLU function outputs the same input value if it is positive; else, it outputs zeros. In most cases, it is used as the activation function of deep neural networks; the activation function is located between layers and removes the linearity of successive inter-layer matrix multiplications. However, we used the ReLU function as the loss function, not the activation function. For ReLU loss, the input is the score of the lost team ($S_L$) minus the score of the won team ($S_W$). \ie, the loss is zero as long as DEP discerns correctly no matter how different both teams' scores are. On the other hand, the loss gets larger by the difference if DEP discerns wrong. The definition of ReLU loss is shown in \cref{eq:reluloss}.

\begin{equation}\label{eq:reluloss}
    ReLU\ Loss =
    \begin{cases}
    S_L - S_W & if\ S_W \leq S_L \\
    0 & if\ S_W > S_L
    \end{cases}
\end{equation}

\subsection{Overall train process}\label{sec:overall}
Our model has ten independent GRU-SLP models since the number of players is 10 in a match. At the beginning of the training, the model randomly initializes its GRU-SLPs parameter values. Then, the training proceeds a match by match. Each GRU-SLPs respectively take input one action sequence of a player in charge and output scores of the actions.

DEP adds up the output scores by teams and discerns the winner by the methods described in \cref{sec:loss_function}. Then DEP calculates the loss by using a loss function. After computing the loss, GRU-SLPs are trained respectively by backpropagation. Finally, the model calculates average parameter values of trained GRU-SLPs and assigns the parameters to each GRU-SLP.

\begin{figure}[ht]
    \centering
    \includegraphics[width=0.9\linewidth]{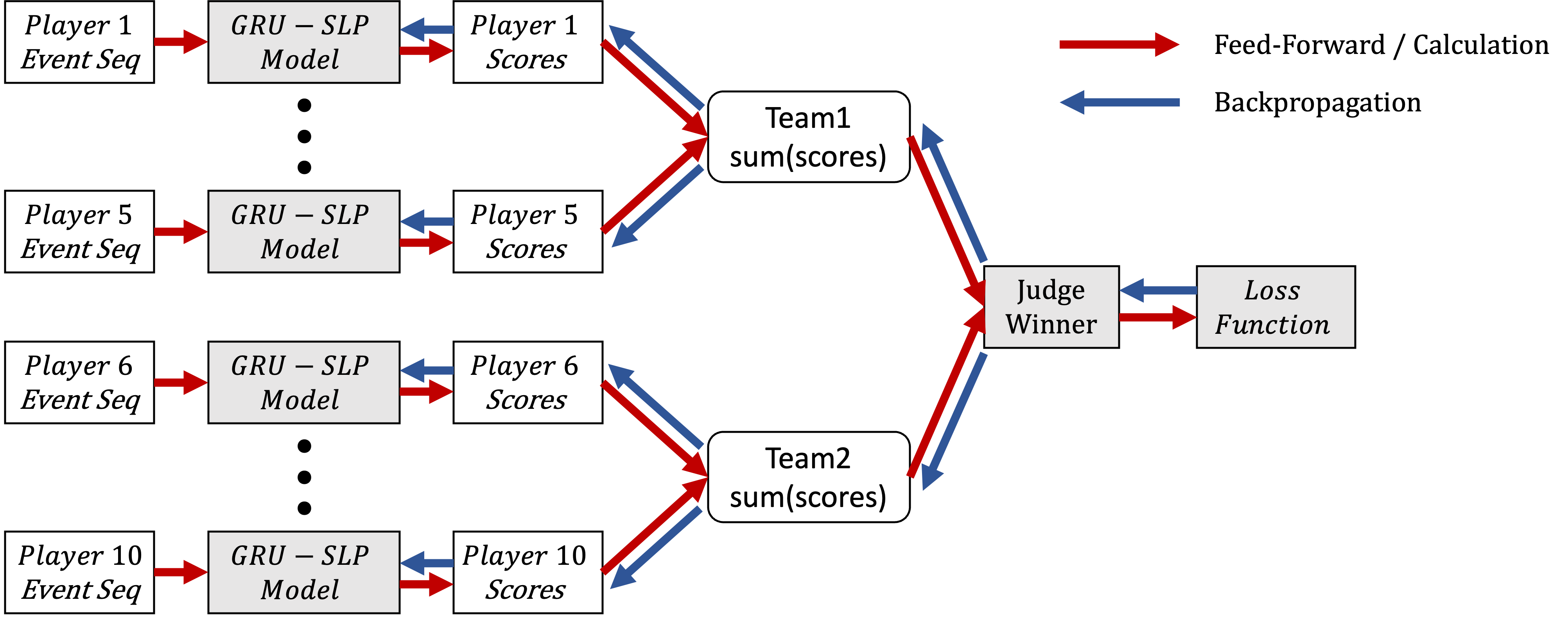}
    \caption{Overall process for a match of the proposed model}
    \label{fig:overall}
\end{figure}

\cref{fig:overall} depicts the overall train process of our model. Furthermore, \cref{algo:parameters} is a pseudo-code that shows how the model trains ten independent GRU-SLPs and assigns the average values of their parameters.

\begin{algorithm}
	\caption{Pseudo codes for training process of proposed model}
	\label{algo:parameters}
	\begin{algorithmic}[1]
	    \State GRU-SLP Model \bb{subModel[10]}
		\For {$match=1,2,\ldots,N$}
			\For {$player=1,2,\ldots,10$}
				\State score[$player$] $\leftarrow$ \bb{subModel[$player$]}(actionSequance[$player$])
			\EndFor
			\State score[Team1] $\leftarrow$ sum(score[1:5]), score[Team2] $\leftarrow$ sum(score[6:10])
			\State prediction $\leftarrow$ PredictWinner(score[Team1], score[Team2])
			\State loss $\leftarrow$ LossFunction(prediction)
			\State Backpropagate and update parameters of \bb{subModels}
			\State \bb{subModels}.parameters $\leftarrow$ average(\bb{subModels}.parameters)
		\EndFor
	\end{algorithmic}
\end{algorithm}

\section{Dataset}\label{sec:dataset}
We collected 245,575 match data of \lol~ by using the RiotAPI, which the company that developed the game (Riot Games) provides \cite{riot_api}. The dataset contains matches of North America, Europe, South Korea, and Japan that were played between May 10 and 17, 2021. We got two JSON files for a match through the API: the match's metadata and the timeline. The metadata has the basic information such as the game's version, chosen champions by each player, total game duration. On the other hand, the timeline data contains significant events' reports and snapshots at 1-minute intervals; an event report has details, including the timestamp and the actor player.

From match data, we set up seven features: timestamp, champion, lane, position, distance, and event. The definition of each feature is elaborated on below.

\begin{description}
    \item[\ii{Timestamp}] is the time point at which the event occurred expressed in milliseconds.
    \item[\ii{Champion}] is the champions chosen by actor players of the event.
    \item[\ii{Lane}] is the chosen lanes of the event's actor player.
    \item[\ii{Position}] represents the location on the map of the actor player when the event occurs.
    \item[\ii{Distance}] depicts how much the actor player is isolated from their teammates.
    \item[\ii{Event}] is the type of the event (such as champion kill and purchase item) and its weight.
\end{description}

\subsection{Champion}
As mentioned in \cref{sec:lol}, A champion belongs to one or more of six roles. We used the \bb{role} as an input feature instead of \bb{champion} since \lol~ has over 140 champions making our model consume more computing power when training. Also, the game company can add or remove champions when they release patches. Therefore, \bb{role} is a more stable feature applicable regardless of the game's version than \bb{champion}.

We created a Champion-Role vector lookup table through \lol~ game data that \ii{Riot Games} officially provides \cite{datadragon} (See \cref{tab:cham-role}). By using the lookup table, \bb{Champion} info of events are converted to one-hot encoded \bb{Role} vector.

\begin{table}[ht]
    \centering
    \begin{tabular}{|c|cccccc|}
        \hline
        Champion   & Assassins & Fighters  & Mages & Marksmen  & Supports  & Tanks \\ \hline\hline
        Annie      & 0         & 0         & 1     & 0         & 0         & 0     \\ \hline
        Kayle      & 0         & 1         & 0     & 0         & 1         & 0     \\ \hline
        Shyvana    & 0         & 1         & 0     & 0         & 0         & 1     \\ \hline
        \multicolumn{7}{|c|}{...}                                                  \\ \hline
        Vayne      & 1         & 0         & 0     & 1         & 0         & 0     \\ \hline
    \end{tabular}
    \vspace{5px}
    \caption{Champion-Role vector lookup table}
    \label{tab:cham-role}
\end{table}

\subsection{Lane}
A player chooses a \bb{Lane} that represents the player's role in a team. There are five official lanes: \ii{Top}, \ii{Mid}, \ii{Bottom}, \ii{Support}, and \ii{Jungle}. The metadata file of a match contains \bb{Lane} info of each player. Like the \bb{Champion} feature, we converted \bb{Lane}s to one-hot vectors too.

\subsection{Position}
\bb{Position} is a specific location on the map of the actor of an event. In a match's timeline data, the location is expressed in x and y coordinates, of which the value is between 0 and 15000. We normalized the values into between 0 and 1 by min-max scaling. Some types of the event do not have \bb{Position} data. Therefore, we presumed the missing \bb{Position} using domain knowledge and a 1-minute interval of snapshots. For example, when an item purchasing event occurs, we can presume the actor's \bb{Position} is (0, 0) or (1, 1) by their team since a player must return to their base to purchase an item.

\subsection{Distance}
\bb{Distance} represents how much the actor player is isolated from their teammates. Since \lol~ is a teamplay game, cooperative and strategic play are crucial to triumph. \bb{Distance} reflects cooperativity and tactics indirectly. For example, when teammates engage in a team fight, a player has to join teammates in most cases. In this situation, a player's \bb{Distance} would be small because the player acts with teammates. On the other hand, a player also can sneak through an enemy's flank to destroy their defense tower when teammates draw the opponents' attention. Opposite to the former situation, the \bb{Distance} would grow. \cref{eq:dist} shows the formula to calculate Distance of player $i$. $D_i$. Digit 1 to digit 4 represent each teammate of player $i$, therefore $D_{nm}$ is the Euclidean distance between player $n$ and $m$.

\begin{equation}\label{eq:dist}
    D_i = \frac{D_{i1}+D_{i2}+D_{i3}+D_{i4}}{D_{i1}+D_{i2}+D_{i3}+D_{i4}+D_{12}+D_{13}+...+D_{34}}
\end{equation}

\subsection{Event}
A match data contains ten different event types representing a player's action, such as purchasing an item and placing a ward. To achieve our goal, we added four new event types derived from original events. The following list shows the original event types in black and additional types in red. We also converted the types of events to one-hot encoded vectors.

\begin{description}
    \item[\ii{ITEM\_PURCHASED}] represents that a player purchased an item.
    \item[\ii{ITEM\_SOLD}] means that a player sold an item they have.
    \item[\ii{ITEM\_DESTROYED}] implies a player consumed a consumable item such as a potion and elixir.
    \item[\ii{SKILL\_LEVEL\_UP}] represents that a player increased their skill level..
    \item[\ii{LEVEL\_UP}] signifies a player's level increased.
    \item[\ii{WARD\_PLACED}] means a player placed a ward on the map for some purpose such as the sight and heals.
    \item[\ii{WARD\_KILL}] implies an action that a player destroys an enemy's ward.
    \item[\ii{CHAMPION\_KILL}] represents that a player killed an enemy's champion.
    \item[\ii{BUILDING\_KILL}] means a player eliminated an opponents' defense tower.
    \item[\ii{ELITE\_MONSTER\_KILL}] represents that a player hunted an elite monster in the jungle area.
    \item[\red{\ii{CHAMPION\_KILL\_ASSIST}}] signifies that a player assisted a teammate to kill an enemy champion.
    \item[\red{\ii{CHAMPION\_KILL\_VICTIM}}] implies that an opponent champion murders a player.
    \item[\red{\ii{BUILDING\_KILL\_ASSIST}}] means that a player assisted a teammate in destroying an enemy defense tower.
    \item[\red{\ii{ELITE\_MONSTER\_KILL\_ASSIST}}] represents that a player assisted a teammate in hunting an elite monster.
\end{description}

\subsection{Event weight}
Each \bb{Event} has a different weight depending on the context, even if the event types are identical. When a player kills an enemy champion, for example, there is a distinction between killing the opponent alone and enlisting the help of teammates. Therefore, we devised formulas to describe the weight of each event, respectively.  We also normalized the weights' values in a range from 0 to 1.
Formulas for the \bb{Event}s are described in \cref{tab:event_weight}:
\begin{table}[ht]
    \centering
    \begin{tabular}{c|c|c}
        Event          & Weight formula   & Description \\ \hline\hline
        \makecell{ITEM\_PURCHASED \\ ITEM\_DESTROYED} & $\frac{item\_purchase\_cost}{highest\_item\_purchase\_cost}$ & \small{\makecell{The numerator is the value of the purchased\\or destroyed item at the event,\\while the denominator is the gold value\\of the most expensive item.}} \\ \hline
        ITEM\_SOLD & $\frac{item\_sell\_cost}{highest\_item\_sell\_cost}$ & \small{\makecell{\ii{highest\_item\_sell\_cost} is the gold value\\that a player receives when they sell\\the most expensive item\\ of all of the game items.}} \\ \hline
        SKILL\_LEVEL\_UP & $\frac{current\_skill\_level}{maximum\_skill\_level}$ &  \small{\makecell{\ii{maximum\_skill\_level} is the maximum level\\of the skill that a player just leveled up\\ at the event, and \ii{current\_skill\_level} is\\the skill level right after the event occurs.}} \\ \hline
        LEVEL\_UP & $\frac{level\_place\_rank}{number\_of\_player}$ & \small{\makecell{ \\The player level ranking within\\ of all of the players decides the event weight.\\{ }}} \\ \hline
        \makecell{WARD\_PLACED \\ WARD\_KILL} & $\frac{ward\_bounty}{highest\_ward\_bounty}$ & \footnotesize{\makecell{\bb{Ward} is a unit that provides a vision to players.\\Each ward has a bounty expressed as gold value,\\and a player can receive it as a reward\\when they destroy the ward.\\ \ii{highest\_ward\_bounty} is the highest bounty\\value of all of the wards.}}\\ \hline
        \makecell{CHAMPION\_KILL \\ CHAMPION\_KILL\_ASSIST \\ CHAMPION\_KILL\_VICTIM} & $\frac{damage\_dealt}{total\_damage\_victim\_received}$ & \footnotesize{\makecell{It represents how much the killer and assists\\contributed to killing the victim each.\\Damages that the killer and assists caused\\to the victim decide the weight.\\It also represents how much the victim\\resisted hard until they got killed.\\Thus, the weight of the victim event is related\\to damages that the victim caused\\to the killers while resists.}} \\ \hline
        \makecell{BUILDING\_KILL \\ BUILDING\_KILL\_ASSIST \\ ELITE\_MONSTER\_KILL \\ ELITE\_MONSTER\_KILL\_ASSIST} & $\frac{1}{number\_of\_involved\_players}$ & \footnotesize{\makecell{The dataset does not have information\\about specific damage dealt or received.\\Therefore, we supposed that every player\\who was involved in the event\\contributed the same as others.}} \\ \hline
    \end{tabular}
    \vspace{5pt}
    \caption{Weight formula by event}
    \label{tab:event_weight}
\end{table}

\subsection{Processed feature vector}
Eventually, a player action is converted to a 30-dimensional vector. Therefore, a player's action sequence becomes a sequence of 30-dimensional vectors. Each vector has 5 ranged values (0 to 1) and 25 one-hot values. \cref{tab:features} shows the features that are converted to vector.

\begin{table}[ht]
    \centering
    \begin{tabular}{|c|c|c|}
        \hline
        \bb{Event}                   & \bb{Description}                     & \bb{Range}    \\ \hline
        Timestamp                    & Time elapsed                         & from 0 to 1   \\
        Mage                         & Champion's role                      & 0 or 1        \\ 
        Fighter                      & Champion's role                      & 0 or 1        \\ 
        Support                      & Champion's role                      & 0 or 1        \\ 
        Tank                         & Champion's role                      & 0 or 1        \\ 
        Assassin                     & Champion's role                      & 0 or 1        \\ 
        Marksman                     & Champion's role                      & 0 or 1        \\ 
        Top                          & Lane was chosen by a player          & 0 or 1        \\ 
        Mid                          & Lane was chosen by a player          & 0 or 1        \\ 
        Bottom                       & Lane was chosen by a player          & 0 or 1        \\
        Utility                      & Lane was chosen by a player          & 0 or 1        \\ 
        Jungle                       & Lane was chosen by a player          & 0 or 1        \\ 
        x\_position                  & x coordinates of a player location   & from 0 to 1   \\ 
        y\_position                  & y coordinates of a player location   & from 0 to 1   \\ 
        distance                     & Isolation from teammates             & from 0 to 1   \\ 
        ITEM\_PURCHASED              & Purchase an item                     & 0 or 1        \\ 
        ITEM\_SOLD                   & Sell an item                         & 0 or 1        \\ 
        ITEM\_DESTROYED              & Destroy an item                      & 0 or 1        \\ 
        SKILL\_LEVEL\_UP             & Player's skill level                 & 0 or 1        \\ 
        LEVEL\_UP                    & Player's level                       & 0 or 1        \\ 
        WARD\_PLACED                 & Placing a ward                       & 0 or 1        \\ 
        WARD\_KILL                   & Kill an enemy ward                   & 0 or 1        \\ 
        CHAMPION\_KILL               & Kill an enemy champion               & 0 or 1        \\ 
        CHAMPION\_KILL\_ASSIST       & Assist in killing an enemy           & 0 or 1        \\ 
        CHAMPION\_KILL\_VICTIM       & Get killed                           & 0 or 1        \\ 
        BUILDING\_KILL               & Destroy an enemy building            & 0 or 1        \\ 
        BUILDING\_KILL\_ASSIST       & Assist in destroying building        & 0 or 1        \\ 
        ELITE\_MONSTER\_KILL         & kill an elite monster                & 0 or 1        \\ 
        ELITE\_MONSTER\_KILL\_ASSIST & Assist in killing monster            & 0 or 1        \\ 
        Event\_Weight                & weight of the event                  & from 0 to 1   \\ \hline
    \end{tabular}
    \vspace{5pt}
    \caption{Converted features}
    \label{tab:features}
\end{table}
\section{Experiment}\label{sec:experiment}
To the best of our knowledge, this study is the first attempt to evaluate the contributions of individual actions; as a result, we used common metrics (KDA, Gold, and Minion kills) as baselines rather than models from previous studies. In addition, we created seven models based on a score embedding methodology to compare performances and determine which method is superior. The tested models are below:

\begin{description}
    \item[Model 1: \ii{ReLU \#1}] Model 1 is a basic version of the model described in \cref{sec:model}. It uses \bb{10 GRU-SLP}s and \bb{ReLU loss} function. Also, $h_0$ of this model (See \cref{fig:gru+slp}) is $(L, 1, N)$-dimension vector that initialized to 0. $L$ is the depth of GRU layers, and $N$ is the length of the hidden dimension of GRU.
    \item[Model 2: \ii{ReLU \#2}] Model 2 is the same as Model 1, but it uses different $h_0$ for won and lost teams. i.e., The winner's $h_0$ is initialized to 1 while the loser is 0. The purpose of the model is to reflect the fact that the consequence of a match (win, lose) occurs just after a player's last action.
    \item[Model 3: \ii{ReLU \#3}] We designed Model 3 to validate our hypothesis that the time-reversed input sequence is adequate to analyze players' actions. Therefore, the action sequences that are input to Model 3 are not time-reversed, but chronological.
    \item[Model 4: \ii{BCE \#1}] Model 4 is a comparison model of Model 1 that uses the Confidence for the discernment method and BCE loss for the loss function.
    \item[Model 5: \ii{BCE \#2}] Model 5 is the same as Model 2 but uses Confidence and BCE loss.
    \item[Model 6: \ii{MLP \#1}] We created Model 6 to validate the GRU-SLP model's effectiveness. Model 6 uses Multi-Layered Perceptrons (MLP) to process each action independently without taking into account the sequential effect. It uses the deterministic discernment method and ReLU loss.
    

    \item[Model 7: \ii{MLP \#2}] Model 7 is the same as Model 6 but uses Confidence and BCE loss for DEP.
\end{description}

\subsection{Model implementation and settings}
We implemented the seven models using \ii{PyTorch} 1.10.0 with CUDA 10.1 \cite{pytorch}. Also, we used \ii{Adam} optimizer \cite{adam} for the models to adjust models' parameters as training progressed. For the GRU-SLPs, the depth of hidden layers is 2, and the dimension of hidden states is 15. Every model was trained in 10 epochs and 0.0001 of the learning rate.

Because of the concern about the overfitting problem, we set test data to consist of 44,575 matches, nearly 20\% of the total in the dataset; the test set represents the unseen data from the trained model. The rest of the dataset, which consists of 200,000 matches, is the trainset. Owing to enough size of the test set, the model would show low performance for the test stage if it is overfitted onto the trainset; it provides confidence that the model that achieves high performance in the test stage is not overfitted. Also, to choose the best-trained model in the training stage, we set the validation stage for each epoch. We extracted the validation data from the trainset, and its size is 5\% (10,000 matches) of the trainset.

\subsection{Discernment accuracy}
First, we checked the discernment accuracy of the models. The accuracy represents how much the models' scores reflect well for the contributions of each action since we trained the models to give high scores for actions that are highly related to winning. \cref{tab:discern_accuracy} shows the discernment accuracy of designed models.

\begin{table}[ht]
    \centering
    \begin{tabular}{c|c|c|c|c}
         & \bb{Accuracy} & \bb{Precision}    & \bb{Recall}   & \bb{F1 score} \\ \hline
        Model 1     & 99.5634\%     & 99.5587\%         & 99.5886\%     & 0.995736     \\
        Model 2     & \bb{100\%}    & \bb{100\%}        & \bb{100\%}    & \bb{1.}    \\
        Model 3     & 99.4800\%     & 99.3884\%         & 99.5971\%   & 0.994927       \\
        Model 4     & 99.4712\%     & 99.5154\%         & 99.4514\%     & 0.994834     \\
        Model 5     & \bb{100\%}    & \bb{100\%}        & \bb{100\%}    & \bb{1.}    \\
        Model 6     & 99.4668\%     & 99.4941\%         & 99.4643\%     & 0.994792     \\
        Model 7     & 99.5063\%     & 99.5285\%         & 99.5071\%     & 0.995178     \\ \hline
        Baseline (KDA)              & 93.2090\% & 93.2399\% & 93.5156\% & 0.933775     \\
        Baseline (Gold)             & 94.5979\% & 95.1655\% & 94.2356\% & 0.946983     \\
        Baseline (Minion kills)     & 65.5381\% & 67.1973\% & 63.8623\% & 0.654874     \\ \hline
    \end{tabular}
    \vspace{5pt}
    \caption{Discernment accuracy, precision, recall and F1 score for each model}
    \label{tab:discern_accuracy}
\end{table}

Model 2 and Model 5 show that the judgment accuracy is 100\% when $h_0$ of the won and the lost team are input differently, regardless of the loss function. It is not surprising that the two models perform flawlessly in the discernment in that those models already knew the ground truth before discerning was executed by receiving the outcomes of matches as input ($h_0$). However, it is only a necessary condition that the model accurately discerns outcomes, not a sufficient condition to achieve this study's purpose. We discuss this topic in \cref{sec:feature-score}. Except for those two models, Model 1 displays the best performance for discerning the winner. In addition, all designed models indicate accuracy exceeding 99\%, while the discernment with common metrics show under 95\% accuracy.

\subsection{Player ranking by models}\label{sec:ranking}
\begin{figure}[hp]
    \centering
    \begin{subfigure}[b]{0.9\linewidth}
        \includegraphics[width=\linewidth]{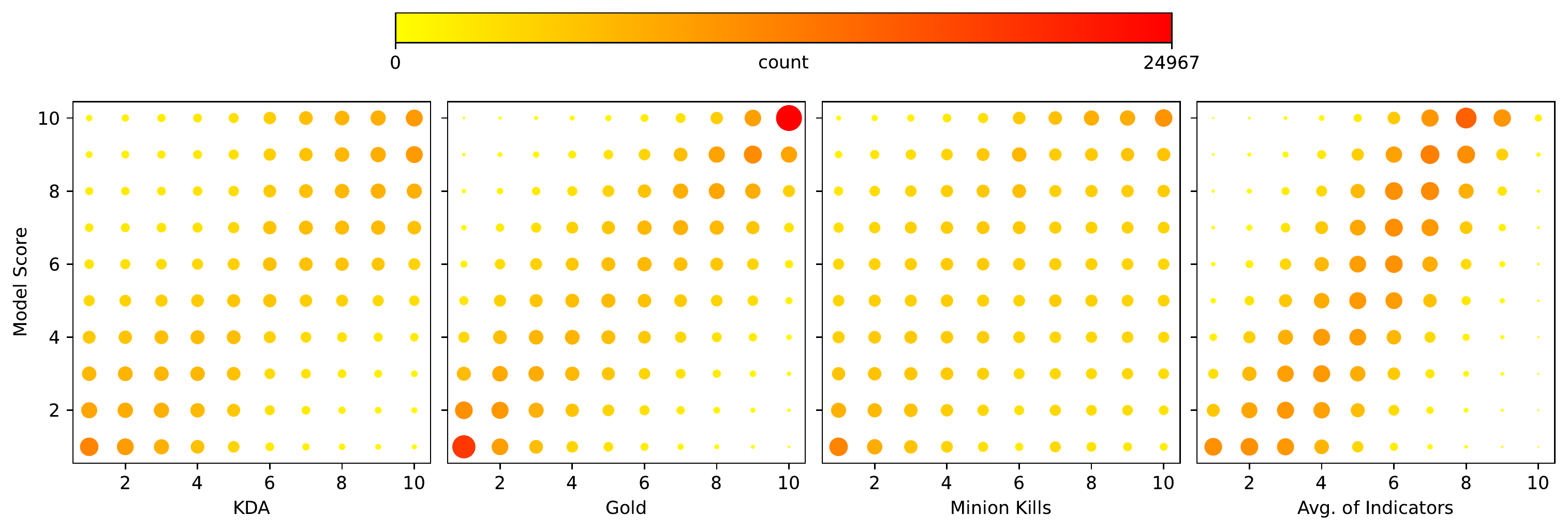}
        \vspace{-10pt}
        \caption{Model 1}
        \label{fig:ranking_model1}
    \end{subfigure}
    \begin{subfigure}[b]{0.9\linewidth}
        \includegraphics[width=\linewidth]{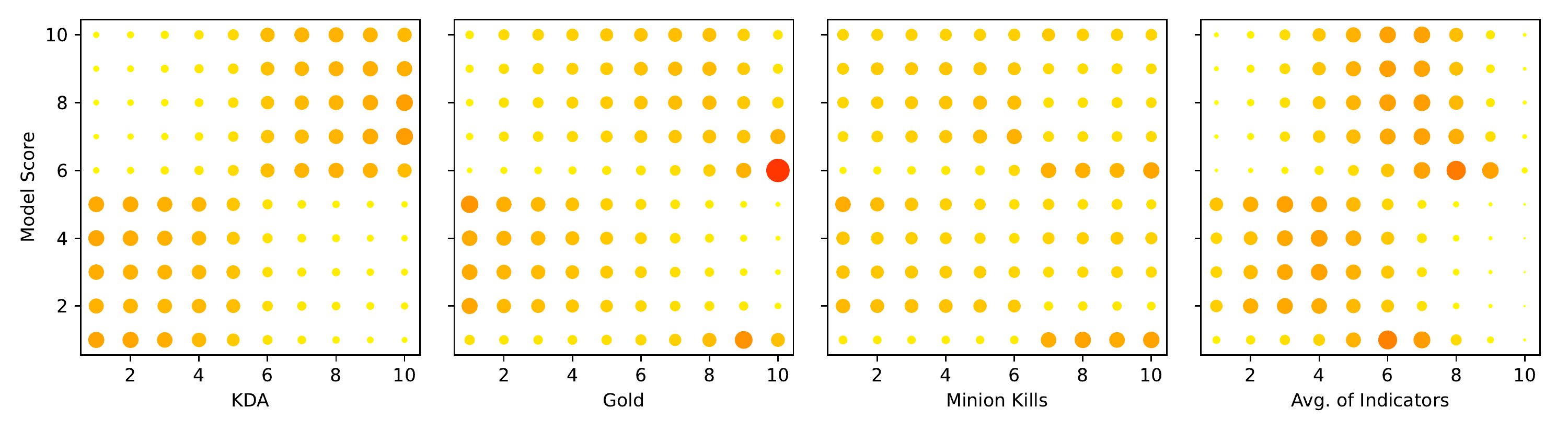}
        \vspace{-10pt}
        \caption{Model 2}
        \label{fig:ranking_model2}
    \end{subfigure}
    \begin{subfigure}[b]{0.9\linewidth}
        \includegraphics[width=\linewidth]{figures/ranking_model5.pdf}
        \vspace{-10pt}
        \caption{Model 5}
        \label{fig:ranking_model5}
    \end{subfigure}
    \hfill
    \begin{subfigure}[b]{0.9\linewidth}
        \includegraphics[width=\linewidth]{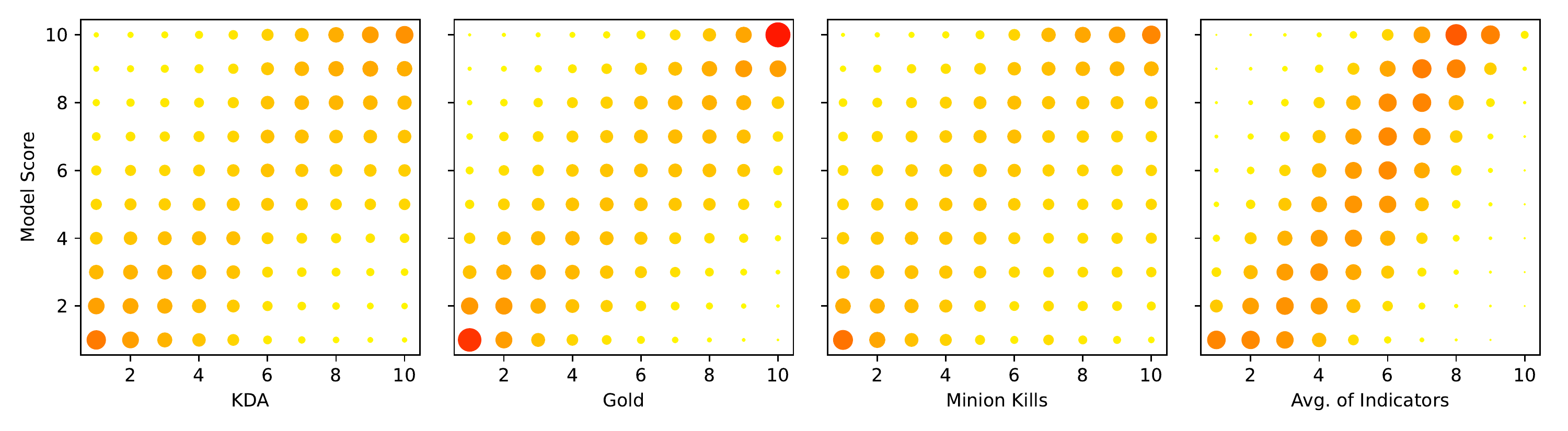}
        \vspace{-10pt}
        \caption{Model 7}
        \label{fig:ranking_model7}
    \end{subfigure}
    \caption{Player ranking counts by model and common metrics}
    \label{fig:rankings}
\end{figure}

Many game communities use common metrics such as KDA, Gold (earn/spent), and Minion kills to assess the individual performance of players. Therefore, we compared the post-match player rankings for common metrics and model scores. To avoid too many graphs on the paper that could confuse, we analyzed the top 4 models on discernment accuracy (\ie, models 1, 2, 5, 7). \cref{fig:rankings} displays the relation between players' model score ranking and typical indicator rankings (KDA, Gold, Minion kills, and the average of them) each. The X-axis denotes players' ranking for typical indicators, and the Y-axis represents players' ranking for the model scores. The color and size of the dots show the number of players ranked in the corresponding rank for the positions in axes. Therefore, if many players have the same rank by both typical indicators and model scores, the circles in the diagonal position become bigger and redder.

All four models showed a positive correlation between the model score and the KDA ranking. However, models 1 and 7 also have a positive correlation with Gold ranking, but models 2 and 5 show a relatively low correlation. In addition, the graphs show that the Minion kills are not highly related to the models' scores compared to other metrics. Furthermore, \cref{fig:rankings} also demonstrate that the ranking average of common metrics has a nearly linear correlation with models' scores. The fact above represents that our models reflect individual performance evaluation by common metrics well. There are some differences; however, we can validate our model with a more detailed analysis since the common metrics are not applicable for every case.

\subsection{Under/overestimated players}\label{sec:under_over}
As mentioned in \cref{sec:lol}, players take charge of one of the five \bb{lanes}, each with a specific function in a team. However, common metrics do not suit every lane of players to estimate their contribution. For example, the general mission of the support lane is to assist other lanes, not to kill an enemy champion. Also, a support lane champion is often targeted in team fights because of their practical supporting skills and quickly murdered because of their fragility. Therefore, a player who takes charge of the support lane usually gets low KDA, and it does not represent that the player contributed less than teammates. A support lane player is frequently underestimated for their contribution by KDA metrics.

We investigated the differences between player rankings by our models and common metrics and analyzed whether the differences were reasonable. For the differences, we defined \ii{underestimated}  (underestimated by common metrics) as a player who ranks over \bb{five places higher in model score than common metrics} and also defined \ii{overestimated} (overestimated by common metrics) as a player who ranks over \bb{five places lower in model score than common metrics}.

\begin{figure}[p]
    \begin{subfigure}[b]{1\linewidth}
        \includegraphics[width=1\linewidth]{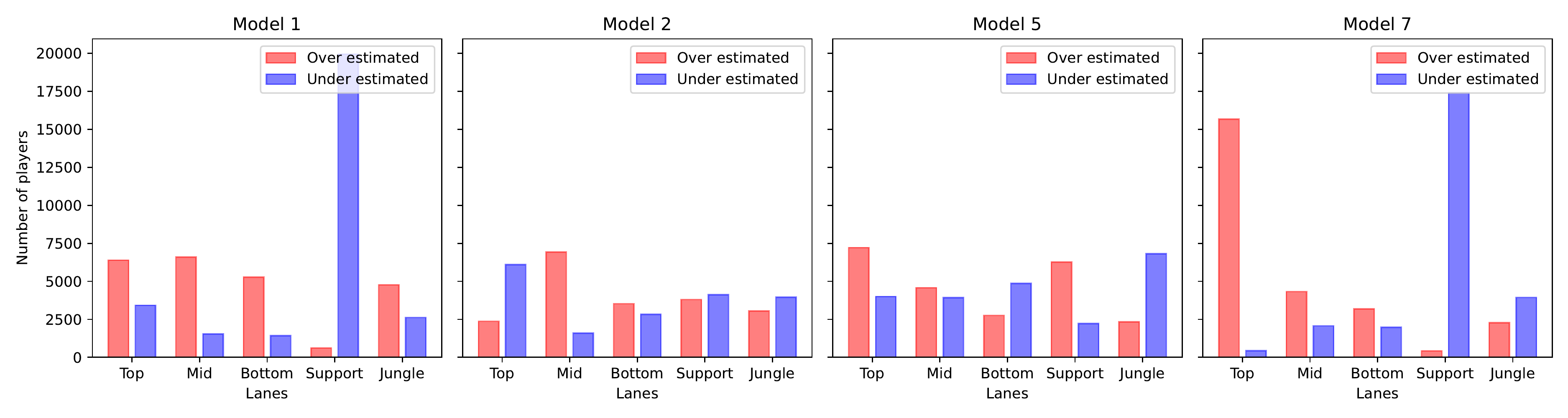}
        \caption{Avg. of common metrics}
        \label{fig:under_over_avg}
    \end{subfigure}
    \begin{subfigure}[b]{1\linewidth}
        \includegraphics[width=1\linewidth]{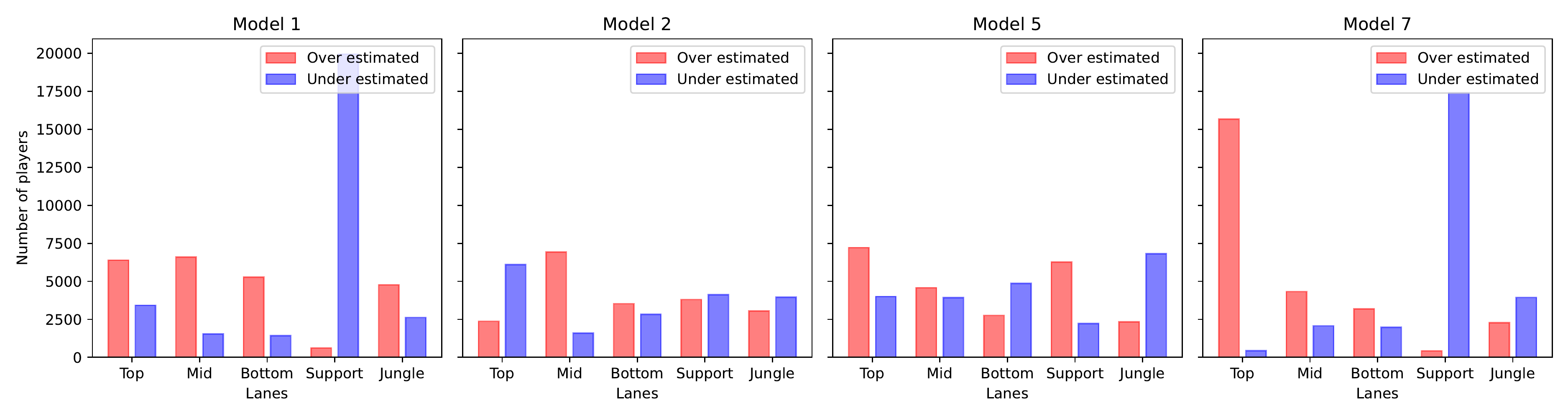}
        \caption{KDA}
        \label{fig:under_over_kda}
    \end{subfigure}
    \begin{subfigure}[b]{1\linewidth}
        \includegraphics[width=1\linewidth]{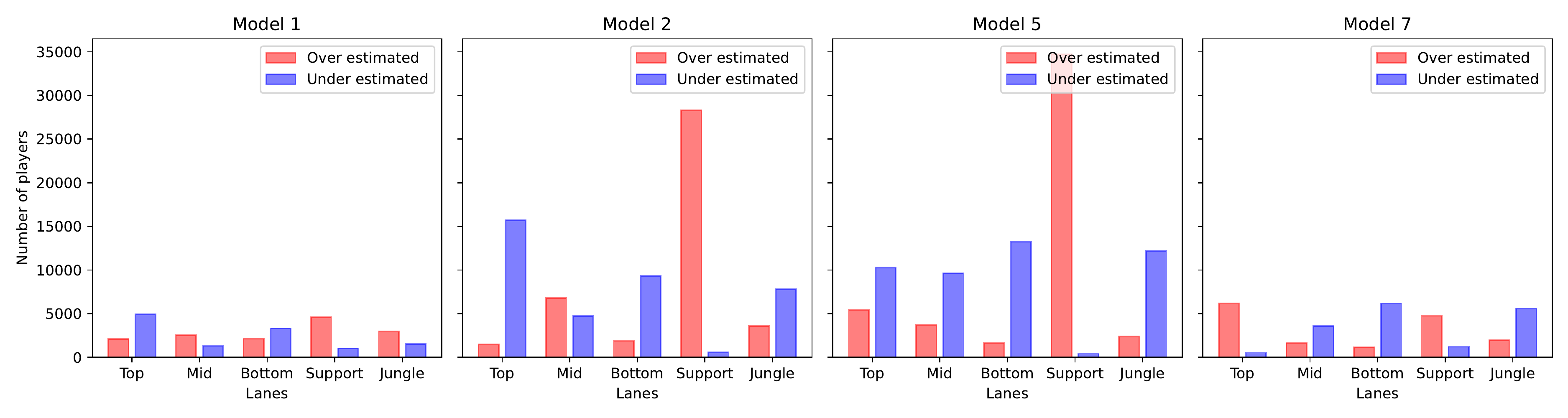}
        \caption{Gold}
        \label{fig:under_over_gold}
    \end{subfigure}
    \hfill
    \begin{subfigure}[b]{1\linewidth}
        \includegraphics[width=1\linewidth]{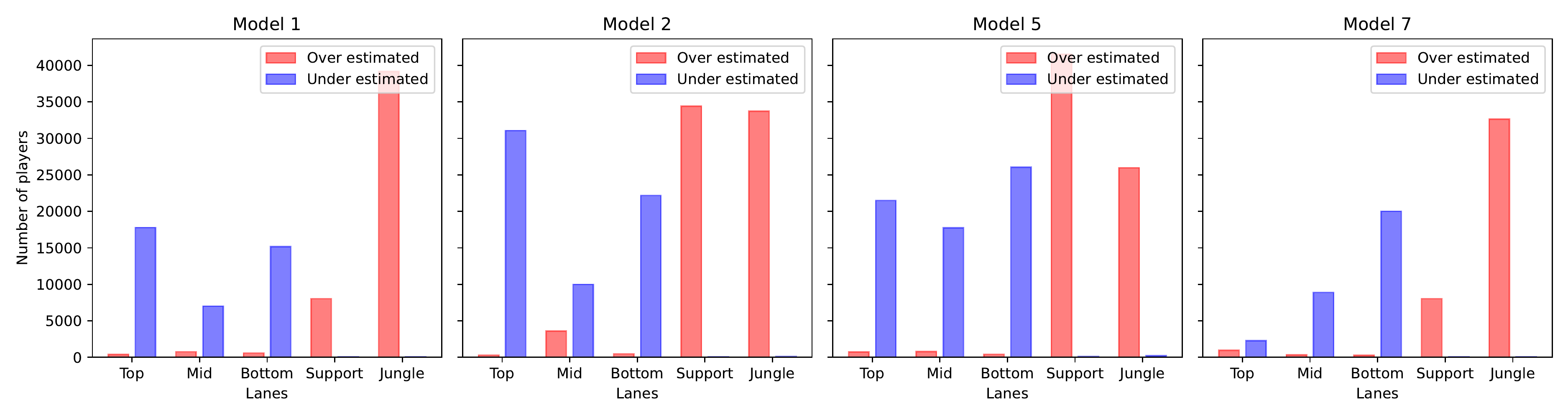}
        \caption{Minion kills}
        \label{fig:under_over_creep}
    \end{subfigure}
    \caption{Number of under/overestimated players by common metrics}
    \label{fig:under_over}
\end{figure}

\cref{fig:under_over} displays the number of under/overestimated players by common metrics. \cref{fig:under_over_avg} to \cref{fig:under_over_creep} are separated statistics by each metric, and each figure consists of separated graphs by models. The red bars of the graphs represent overestimated players, and the blue bars represent underestimated players. \cref{fig:under_over_kda} shows the KDA underestimates many support lane players compared with model 1 and model 7. It is reasonable that the models give a higher evaluation to support lane players than the KDA metrics, as mentioned above about the support lane players' missions.

On the other hand, every model gives low scores to players who took the support and jungle lane. For support lane players, it is a general strategy that the support player concedes the advantage of killing minions to teammates. If a support player monopolizes advantage from killing minions than teammates, teammates' growth who have a role to battle with enemies will be slow, then the possibility of loss becomes higher. Meanwhile, the mission of a jungle lane player is to clear field monsters and gather benefits from them. If a jungle player spends too much time killing minions, the enemy jungle player can monopolize the field monsters' benefit and make it slow that teammates grow. Therefore, the differences of our models' ranking and the minion kill count ranking are rational.

\subsection{Feature-score correlation analysis}\label{sec:feature-score}
\begin{figure}[ht]
    \centering
    \begin{subfigure}[b]{0.47\linewidth}
        \includegraphics[width=\linewidth]{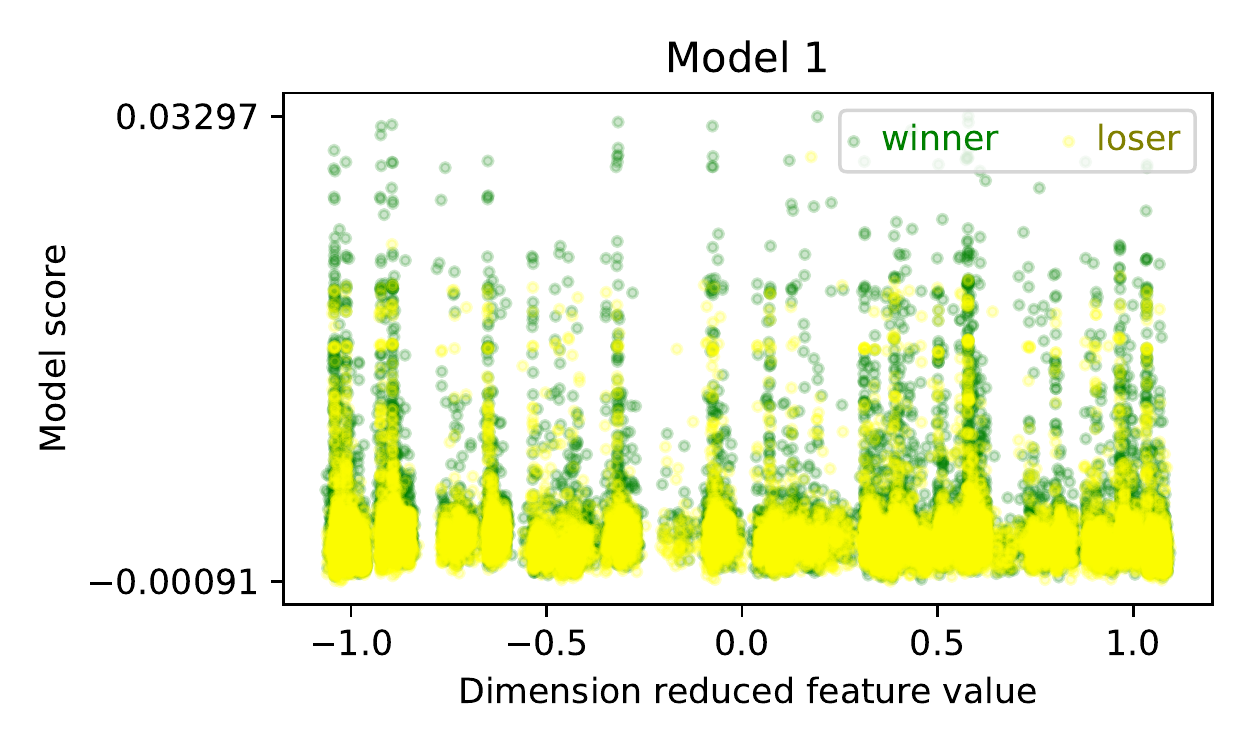}
        \caption{Model 1}
        \label{fig:pca_model1}
    \end{subfigure}
    \hfill
    \begin{subfigure}[b]{0.47\linewidth}
        \includegraphics[width=\linewidth]{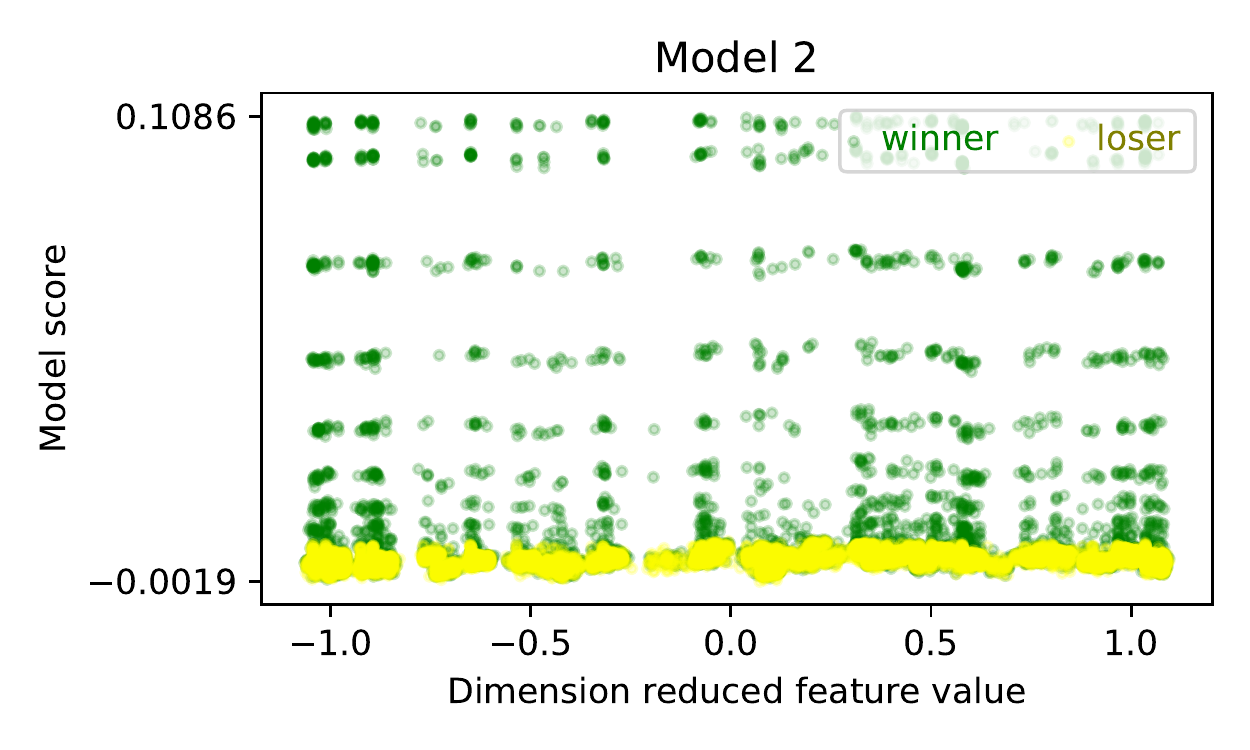}
        \caption{Model 2}
        \label{fig:pca_model2}
    \end{subfigure}
    
    \begin{subfigure}[b]{0.47\linewidth}
        \includegraphics[width=\linewidth]{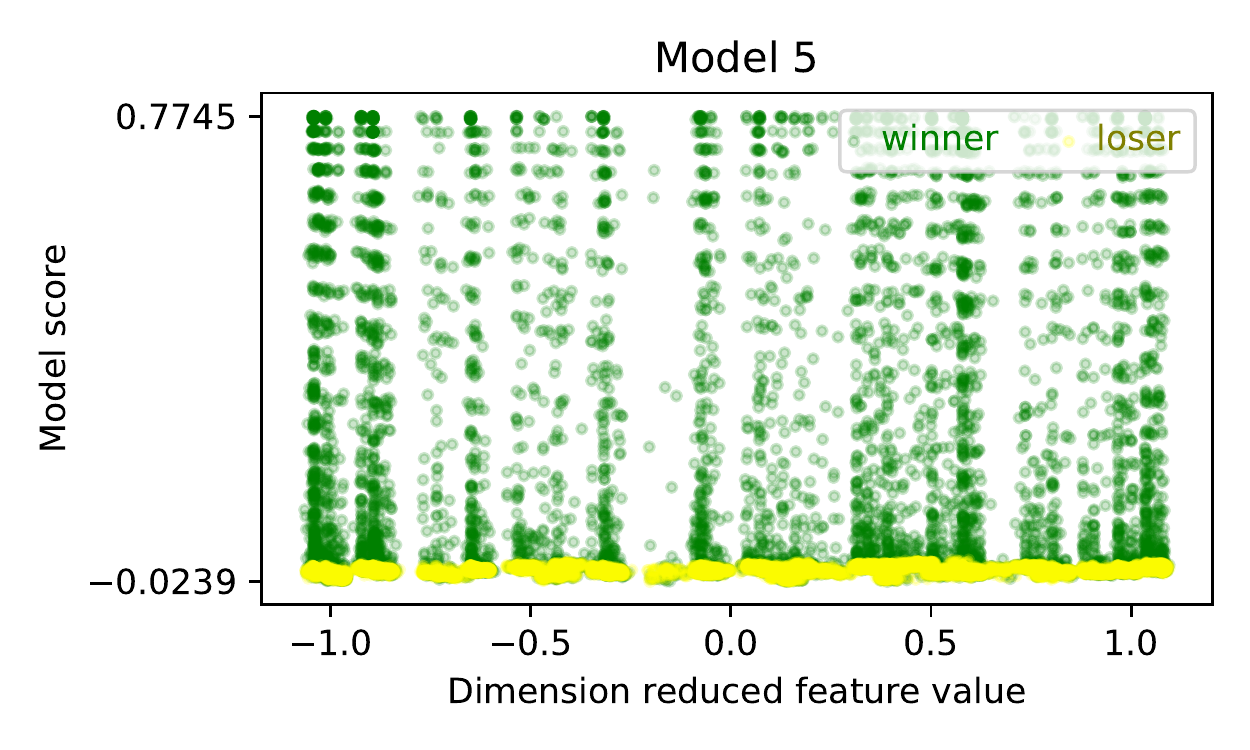}
        \caption{Model 5}
        \label{fig:pca_model5}
    \end{subfigure}
    \hfill
    \begin{subfigure}[b]{0.47\linewidth}
        \includegraphics[width=\linewidth]{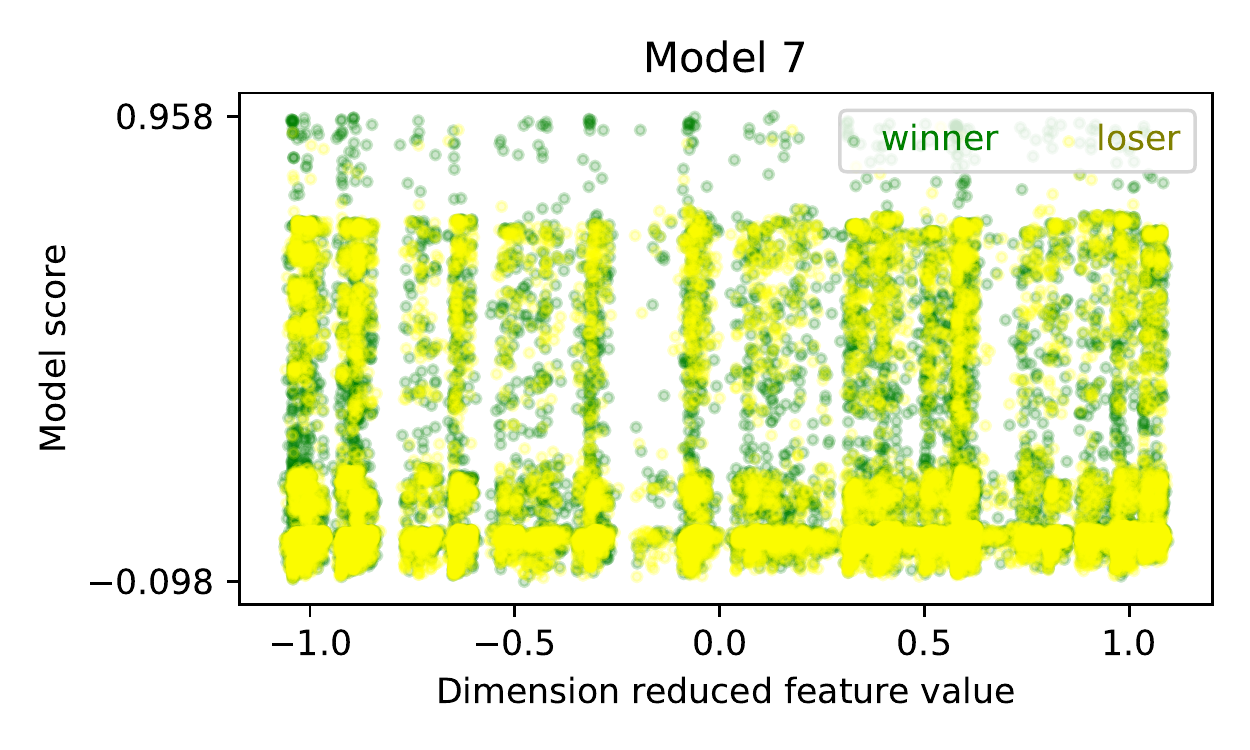}
        \caption{Model 7}
        \label{fig:pca_model7}
    \end{subfigure}
    \caption{Relationship between models' score and features}
    \label{fig:pca_relation}
\end{figure}

To achieve our goal that measures a player's contribution regardless of whether they won or lost, the embedding model has to give similar scores to the same actions in a similar context. Therefore, we examined the score distribution by the action features. To visualize the relationship between the model scores and all features at once, we reduced the feature dimension to 1 using Principal Component Analysis (PCA), then drew graphs displaying the relationship between model scores and dimension-reduced features (See \cref{fig:pca_relation}). We can consider that actions with the same dimension-reduced feature value have similar worthiness in a match context. Therefore, to fulfill the goal of this study, the actions with the same dimension-reduced feature value have to get similar scores from the trained model without bias by the matches' outcomes.

In \cref{fig:pca_relation}, the green dots represent the winners' scores according to the dimension-reduced feature values, and the yellows represent the losers' scores. If the model fulfills our purpose, the clusters of green and yellow dots will ting similar shapes, representing that the model scores actions by their value, not affected by the matches' outcomes. On the other hand, if the green and yellow clusters ting significantly different shapes, we can consider that the model's scoring function is heavily affected by the outcomes of the matches. As we can see from the graph, Model 2 and Model 5 give scores significantly different for winners and losers. Therefore, those two models do not suit our purpose despite the discerning accuracy being 100\% since they evaluate a player's actions considering the outcome of a match, not the actions and contexts themselves. On the other hand, Model 1 and 7 give similar scores for similar dimension-reduced feature values. Therefore, although their discerning accuracy is relatively lower than Model 1 and 5, \cref{fig:pca_relation} shows those two models are more suited to the purpose of this study.

\section{Discussion}\label{sec:discussion}
In the experiment section (\cref{sec:experiment}), we looked at the winner-discernment accuracy of the models to see if they accurately reflect a player's entire contribution to team’s win. The results (\cref{tab:discern_accuracy}) show that our algorithms accurately discern the outcome of a game. Then we compared our models to common metrics to see if they estimate a player's performance properly. Despite minor differences, the models' outputs are comparable to common metrics. Furthermore, as we saw in \cref{sec:under_over}, the discrepancies between our models (particularly Model 1 and Model 7) are rational and explainable. Finally, we looked at the feature-model score correlation to see if the models gave similar scores to actions with similar feature values (worthiness in a match context). \cref{fig:pca_relation} depicts that Model 2 and Model 5 give significantly different scores to the won player's action and the lost player's action, despite their similar feature value; in contrast, models 1 and 7 give similar scores to similar actions regardless of the match's outcome. The last analysis shows that Model 1 and Model 7 suit our research purpose more than Model 2 and Model 5 despite their relatively low discerning accuracy.

There are limitations to this study. First, the RiotAPI that we used to construct the dataset provides limited types of action. Our dataset does not have a player's decision and specific acts in detail, such as skill usage and giving damages to the defense tower. Therefore, our approach is not tested for a dataset that contains very detailed action information. It will be our future work to validate our model by collecting a more detailed dataset; it can be done with advanced data collection methods such as replay video analysis. Also, our model cannot be used for real-time services such as e-sports commentary because we designed it for post-match evaluation.

Our models' strength, despite their limitations, is their generality and maintainability. Because the actions used for scoring are generic enough, the trained models do not need to retrain every time the game is updated with adding a new champion and item or rebalancing the game system.

Furthermore, despite the actions and attributes for \lol~do not precisely fit other games, we can use the fundamental structure of our model (which is the pivotal contribution of this paper) for most games, including different genres such as First-Person Shooter (FPS). The only task developers need to finish before using our model is defining actions and the actions' attributes representing their game well.
  
\section{Conclusion}\label{sec:conclusion}
Estimating individual performance for fair reward and analyzing past behaviors to improve a player's skill and team-level tactics are fascinating subjects of team-based competitive sport study. This study introduced an embedding approach to score player's in-game actions by how much they contributed to winning. The idea of our approach came from the word-embedding technique, NNLM. This approach allows quantitative evaluations of a player's respective actions, which were not available from common metrics and previous studies.

With quantitative values of players' contribution to victory, MOBA games can use our model in many ways. First, our model can adjust the amount of MMR that increases or decreases according to match outcome, alleviating the side-effects of the team/result-oriented MMR systems (The exact amount of MMR that our model adjusts could vary from game to game, according to the MMR formula and the developers' intention). The second application is to display contribution scores on the games' season leaderboard or to inform players at the ends of matches. As a result, the games can honor players who have made significant contributions to their teams' accomplishments but have been overlooked by traditional performance criteria. This can avoid situations where some players unreasonably blame their teammates for low KDA or gold despite their enough contributions. Finally, the contribution scores can be utilized to maintain professional players' records in the professional e-sports business. Furthermore, our model's applicability is not limited to MOBAs; the GRU-SLP and DEP structures can be used in various genres.


\end{document}